# UNSUPERVISED MACHINE LEARNING TO ANALYSE CITY LOGISTICS THROUGH TWITTER

Simon Tamayo, Centre for Robotics - MINES ParisTech PSL, France
François Combes, IFSTTAR/AME/SPLOTT, France
Arthur Gaudron, Centre for Robotics - MINES ParisTech PSL, France

**KEYWORDS:** City Logistics, Machine Learning, Natural Language Processing, Social Media Mining, Sentiment Analysis.

## ABSTRACT

City Logistics is characterized by multiple stakeholders that often have different views of such a complex system. From a public policy perspective, identifying stakeholders, issues and trends is a daunting challenge, only partially addressed by traditional observation systems. Nowadays, social media is one of the biggest channels of public expression and is often used to communicate opinions and content related to City Logistics. The idea of this research is that analysing social media content could help in understanding the public perception of City logistics. This paper offers a methodology for collecting content from Twitter and implementing Machine Learning techniques (Unsupervised Learning and Natural Language Processing), to perform content and sentiment analysis. The proposed methodology is applied to more than 110 000 tweets containing City Logistics key-terms. Results allowed the building of an Interest Map of concepts and a Sentiment Analysis to determine if City Logistics entries are positive, negative or neutral.

## INTRODUCTION

Social media is defined as mobile and web-based technologies to create interactive platforms via which individuals and communities share, co-create, discuss, and modify user-generated content (Kietzmann et al. 2011). The main social networks that we know today were created in the mid- 2000s (Boyd & Ellison 2007) and have become an important part of our society, as they have given web users the means for sharing content about different topics. Such content can be analysed in order to extract valuable information, this is known as *social* *media mining*, that is, the process of representing, analysing, and extracting actionable patterns from data collected from social media (Gundecha & Liu 2012).

City Logistics could profit greatly from social media mining, as it deals with different stakeholders, whose engagement is key to enabling and facilitating the implementation of measures (Holguín-veras et al. 2018). This paper proposes an implementation of Machine Learning techniques in order to perform social media mining about City Logistics using Twitter data. Twitter, with 326 million monthly active users and over 500 million messages per day in 2018 (Twitter 2018), has become an important source of information for analysing opinions and sentiments. This research is inspired by the works of Olson et Al. (Olson & Neal 2015) and Kruchten (Kruchten 2014) who proposed an analysis of the preferences of the users posting on the website *Reddit*. The theory behind this type of study is that of Latent Semantic Analysis (LSA): a technique in Natural Language Processing to analyse relationships between a set of documents and the terms they contain. The underlying idea of LSA is that the aggregate of all the contexts in which a given word does or does not appear provides a set of mutual constraints that largely determines the similarity in meaning of words and sets of words to each other (Landauer et al. 1998). In other words, LSA assumes that words that are close in meaning will occur in similar pieces of text. In recent years, the development of open and accessible Machine Learning libraries, such as scikit-learn (Pedregosa et al. 2011), has allowed the implementation of these techniques to many domains, such as medicine (Allen et al. 2016; Oscar et al. 2017) and politics (Tsou et al. 2013).

This paper applies these concepts to City Logistics and examines how they can contribute to its observation and analysis. Specifically, it addresses the following research questions: What are the most frequently shared concepts about City Logistics? How are these concepts organized? Are there some under-represented issues? What has been the evolution of City Logistics on Twitter? Is the perception of City Logistics positive, neutral or negative? Has this perception changed over time? What has been the evolution and perception of key subjects such as *Low Emission Zones* and *Urban Distribution Centres*?

In order to answer these questions, two Machine Learning techniques are used in the proposed analysis: (i) dimensionality reduction and (ii) clustering. Dimensionality reduction is the process of reducing the number of variables under consideration by obtaining a set of principal variables (Roweis & Saul 2000). Clustering is the process of grouping a set of objects in such a way that objects in the same group are more similar in some particular manner to each other than to those in other groups (Ghuman 2016).

The paper proceeds as follows: firstly, the motivation for using social media mining as a source of information about City Logistics is discussed. Then, the corpus constitution and data analysis techniques are described. Next, the findings are presented and discussed.

# MOTIVATION

City Logistics is an essential economic function of urban areas. The role of logistics is to make goods (and services) available to consumers efficiently, both in terms of costs and customer service. However, City Logistics also raises a number of policy issues regarding transport (congestion, safety, noise, local pollution, etc.), land use (logistical lock-in, logistic sprawl, etc.), economics (firm performance, attractiveness, precarious work, etc.) and climate change. Stakeholders are varied (shippers, receivers, carriers, consumers, inhabitants, national and local governments, etc.), therefore, City Logistics policymaking is complex as it requires diagnosis and analysis, thus observation. But the observation of City Logistics is very challenging for several reasons: (1) Relevant data is often strategic for companies, thus secret or expensive; (2) City logistics is a transversal issue: it is relevant to several distinct institutions. However, each institution is only concerned with part of the system. Why would they pay for information that is not directly relevant for them? (3) City Logistics is often low on political agendas: it is not easy for policymakers to link it *as a whole* to the concerns of their voters. And (4) City Logistics is a fast-changing system: trends such as just-in-time, e-commerce, omni-channel, sharing economy, etc., have complex, cascading consequences.

Observation of City Logistics traditionally combines statistics (Serouge et al. 2014) and qualitative observation, typically relying on professional, technical or academic communities. This is sometimes organised in the form of logistics observatories (OECD/ITF 2016) or of hybrid approaches, such as statistics on opinions (The World Bank 2016). These approaches have qualities but also limitations. Quantitative surveys are rigorous and transferable but provide little insight outside their domain of validity, and academic and professional groups have limited information processing capabilities and can be subject to significant biases. Social media mining is therefore an opportunity to complete these protocols. This paper's motivation is to explore to what extent.

## METHODOLOGY

The proposed methodology for analysing City Logistics content is shown in Figure 1. Data collection was performed by scraping the Twitter website with the search terms "City Logistics", "Last Mile Logistics", "Urban Logistics" and "Urban Freight". The collected entries (i.e. tweets) were filtered in order to erase repeated entries. The text cleaning and lemmatization consists in removing undesired content from the data (such as links, symbols and linking words) and then lemmatizing the text inputs. Lemmatization is the process of grouping several forms of a word together so they can be analysed as a single item. For example, the verb "to contribute" may appear as "contributed", "contributes", "contributing", etc. The base form "contribute" (i.e. the one in the dictionary) is called the lemma of the word.

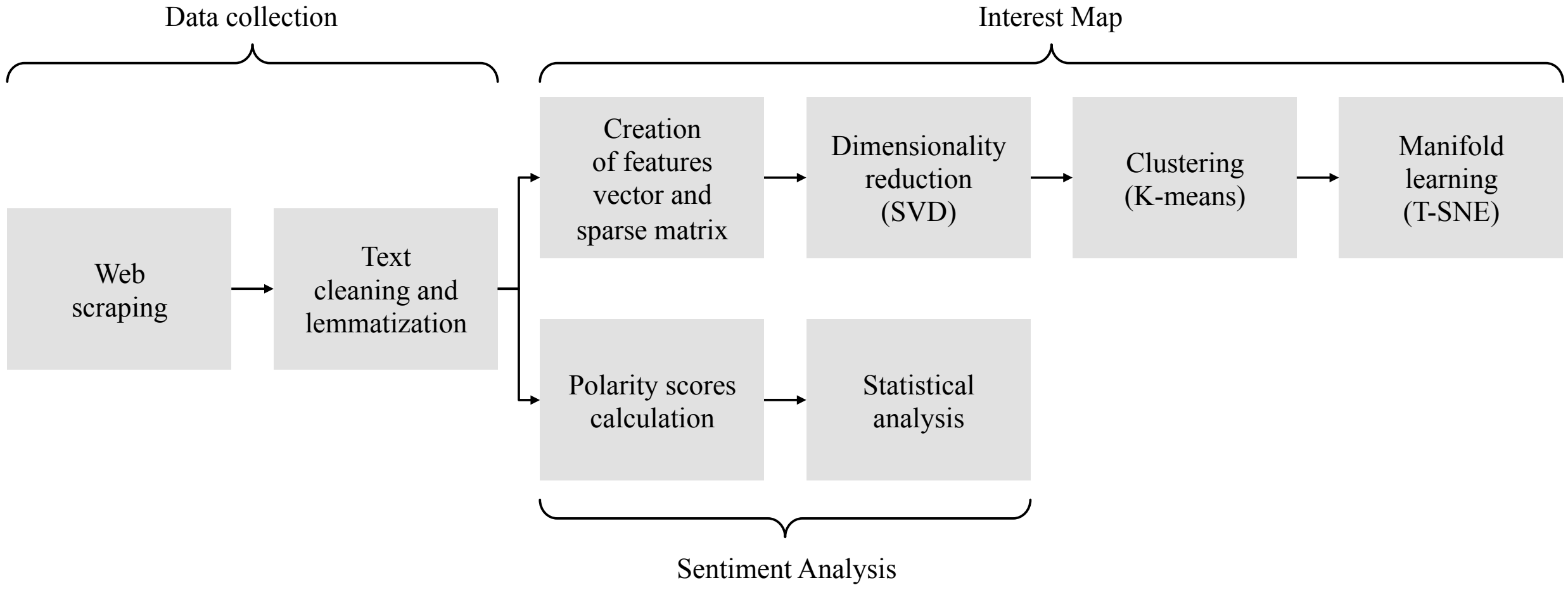

Figure 1. Methodology for analysing Twitter content

In the first part of the analysis we built an interest map of features by performing 4 steps: (1) Input content is transformed into a features vector in which the lemmas are grouped by n-grams (sets of 1, 2 or 3 words), then this vector is used to build a sparse matrix, which is a binary matrix that indicates if each feature is present in each entry. The sparse matrix has a very large number of dimensions and is almost empty. Next, dimensionality reduction (2) is performed. We used Truncated Singular Value Decomposition to reduce the number of dimensions. SVD was preferred over Principal Components Analysis, because it can work with sparse matrices more efficiently (Pedregosa et al. 2011). The resulting matrix is denser and has continuous values. We then applied the K-Means algorithm to the data (3) in order to group features that are "close" in terms of user interest. Finally, a manifold learning algorithm was applied (4) to obtain a two-dimensional result. The used algorithm was t-Distributed Stochastic Neighbour Embedding, which allowed to reveal data that lie in multiple different manifolds or clusters (Pedregosa et al. 2011; Van Der Maaten & Hinton 2008). The resulting interest map is a 2D scatter plot shown in Figure 3.

The second part of the methodology performed sentiment analysis on the inputs. Sentiment analysis is the procedure in which information is extracted from the opinions, appraisals and emotions of people in regards to entities, events and their attributes (Unnisa et al. 2016). In this research, we were interested in finding if the tweets related to City Logistics had positive, negative or neutral sentiments. This analysis was performed using the Nltk library (Bird et al. 2009). The first step of sentiment analysis consists in calculating the polarity score (negative vs. positive) of each input document, this was done with VADER (Valence Aware Dictionary and sentiment Reasoner), a rule-based sentiment intensity analyser (Hutto & Gilbert 2014). The second step consists in computing traditional statistics.

## APPLICATION AND FINDINGS

The proposed methodology was applied to a set of 111 265 tweets containing City Logistics key-terms posted between 2007 and 2018. The evolution of the number of tweets is shown in Figure 2. It is important to highlight that Twitter was created in 2006 but started becoming popular in 2007; this explains the reduced number or occurrences at the beginning of the timeline.

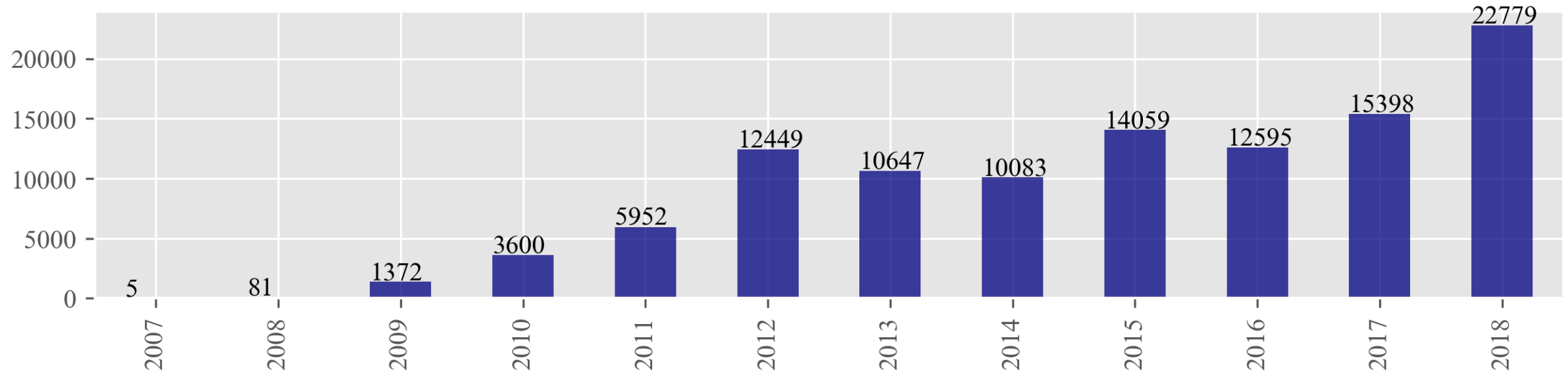

Figure 2. Number of tweets per year

### Vocabulary and frequently used terms

This collection of data allowed the identification of the most used vocabulary related to City Logistics. It was found that the term *City Logistics* is preferred to *Last-mile Logistics*, *Urban Logistics* and *Urban Freight*. Table 1 presents the number of entries obtained of each key-term.

Table 1. Number of entries per key-term

| **Key-term** | *City Logistics* | *Last-mile Logistics* | *Urban Logistics* | *Urban Freight* |
|---|---|---|---|---|
| **Nb. of tweets** | 73 802 (~66%) | 21 219 (~19%) | 9 721 (~9%) | 6 523 (~6%) |

After grouping all entries in a single corpus, statistical analysis was performed to find the most frequent n-grams. In computational linguistics, an n-gram is a contiguous sequence of n items from a given text (Broder et al. 1997). N-grams of sizes 1, 2 and 3 are referred to as unigram, bigram and trigram respectively. Table 2 shows the top 5 unigrams, bigrams and trigrams in the analysed corpus. Unsurprisingly, the researched key-terms appear in the top five but other than that, the most frequent n-grams are those related to *Employment*, such as *job* and *cdl (commercial driver's license)*. It is interesting to find *Kansas City* in the top-5 bigrams. Kansas serves as a key transit point for commerce in the U.S. (Diaz-Camacho 2017) which clearly translates into an important social media activity related to City Logistics.

Table 2. Top5 n-grams (n=[1,2,3])

| **Top 5 unigrams** | | **Top 5 bigrams** | | **Top 5 trigrams** | |
|---|---|---|---|---|---|
| *Job* | 32 489 | *Last mile* | 14 867 | *Last mile logistics* | 7 077 |
| *Urban* | 15 874 | *Mile logistics* | 7 182 | *Salt Lake City* | 5 143 |
| *Mile* | 15 601 | *Kansas City* | 6 909 | *Job cdl logistics* | 4 432 |
| *Last* | 15 382 | *City logistics* | 5 864 | *Cdl logistics needed* | 4 432 |
| *Cdl* | 10 386 | *Oklahoma City* | 5 666 | *Lake City Utah* | 4 087 |

**Interest map**

Many representations can be proposed to display the concepts in the corpus. In this section we present the interest map generated with the methodology in Figure 1. The interest map shown in Figure 3 contains the 10 000 most frequent unigrams, bigrams and trigrams.

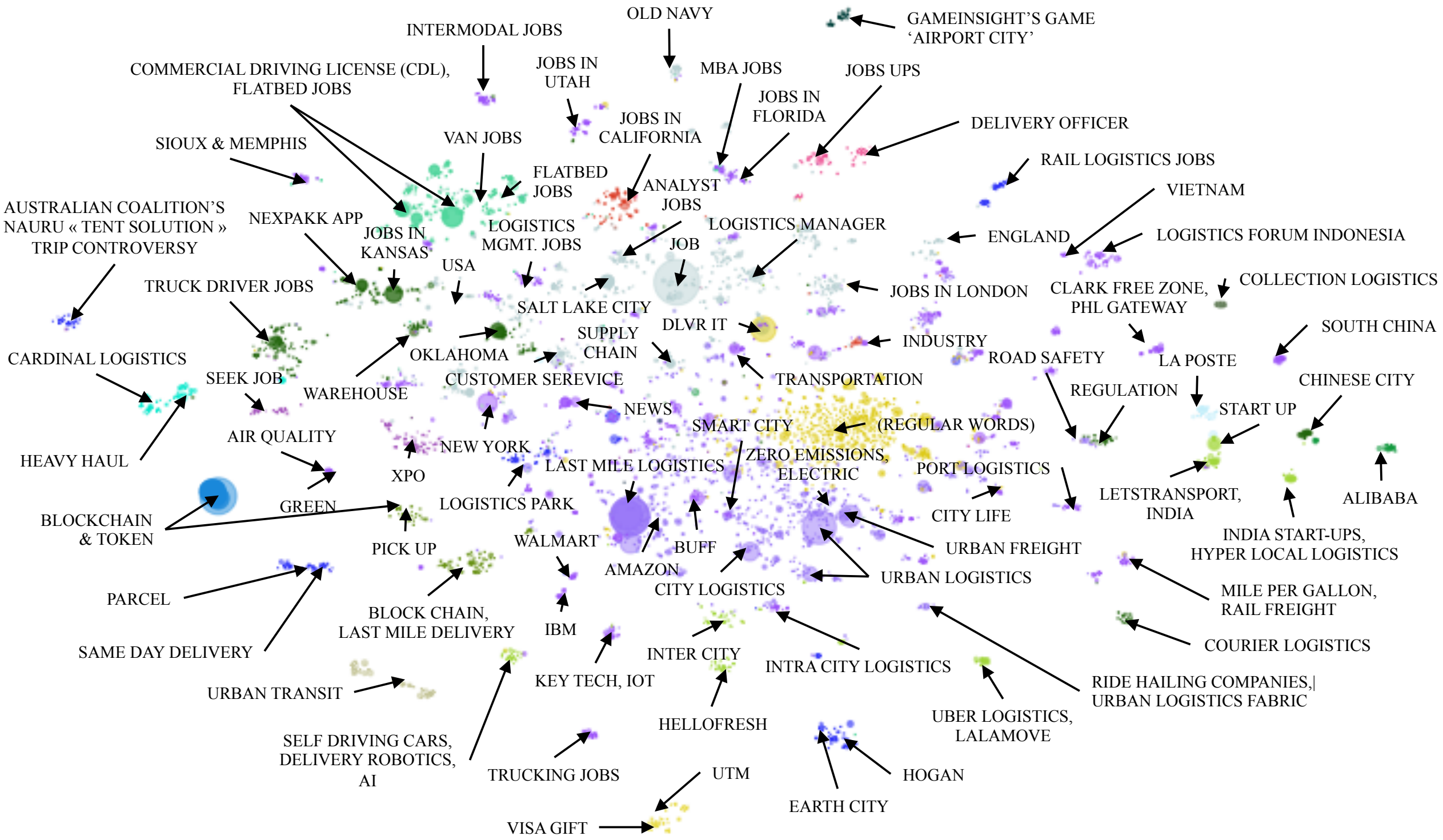

Figure 3. Interest map of City Logistic concepts in Twitter

Each dot represents an n-gram and its size is proportional to its prevalence. In order to explore this visualization in detail, the reader is encouraged to visit the interactive version of the map available at: *http://chairelogistiqueurbaine.fr/2018/10/15/1072*. The algorithms used to create this map assume that concepts that are close in meaning will occur in similar entries of text. Therefore, the resulting visualization implies that concepts represented by nearby points are similar (i.e. they are often present in the same entries) and distant points represent dissimilar objects (i.e. rarely seen together).

## Sentiment analysis

The results of the sentiment analysis reveal that the overall content of the tweets related to City Logistics is 45.7% neutral, 46.7% positive and 7.6% negative, as shown in Figure 4. It should be noted that this distribution has not been constant in time. For example, the proportion of positive content has increased in recent years (43%, 49% and 68% in 2016, 2017 and 2018 respectively).

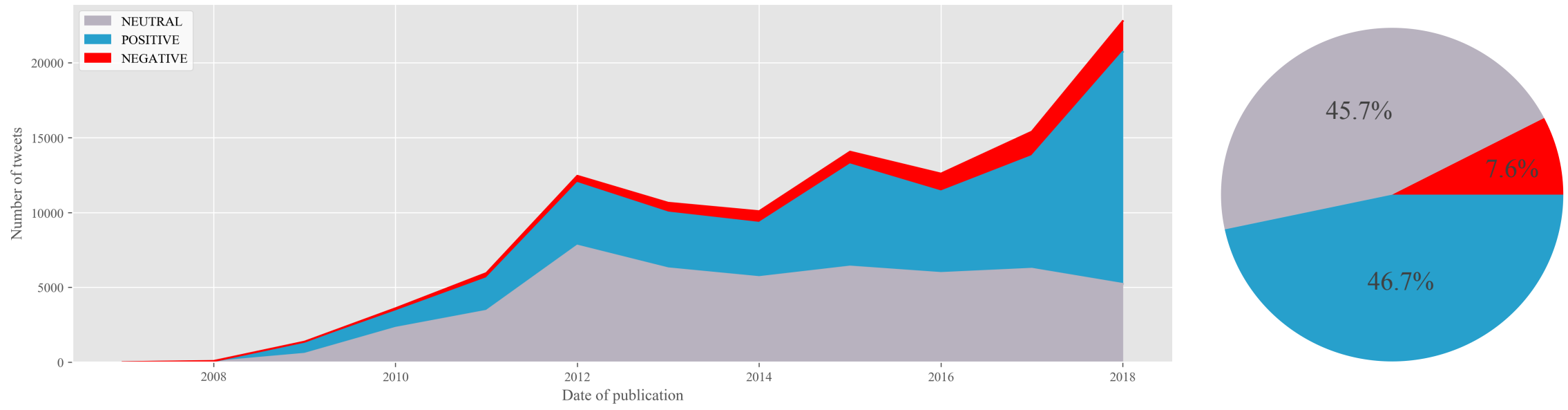

Figure 4. Sentiment analysis of City Logistics tweets

# DISCUSSION

The interest map (cf. Figure 3) illustrates the very large variety of topics of City Logistics tweets. Each of the numerous small groups in the map can be associated to a theme. In Figure 5 below, points are grouped into five aggregate categories, namely: *jobs* (mainly job offers ranging from truck driver to supply chain manager); *technologies* (blockchain, self-driving cars, IoT); *start-ups or innovative firms* (Uber logistics, Letstransport, La Poste); *Asia* (Asian cities, locations and firms); and *core trends and issues*. Some groups are left alone: they are actually not relevant to City Logistics. They are located on the graph's periphery.

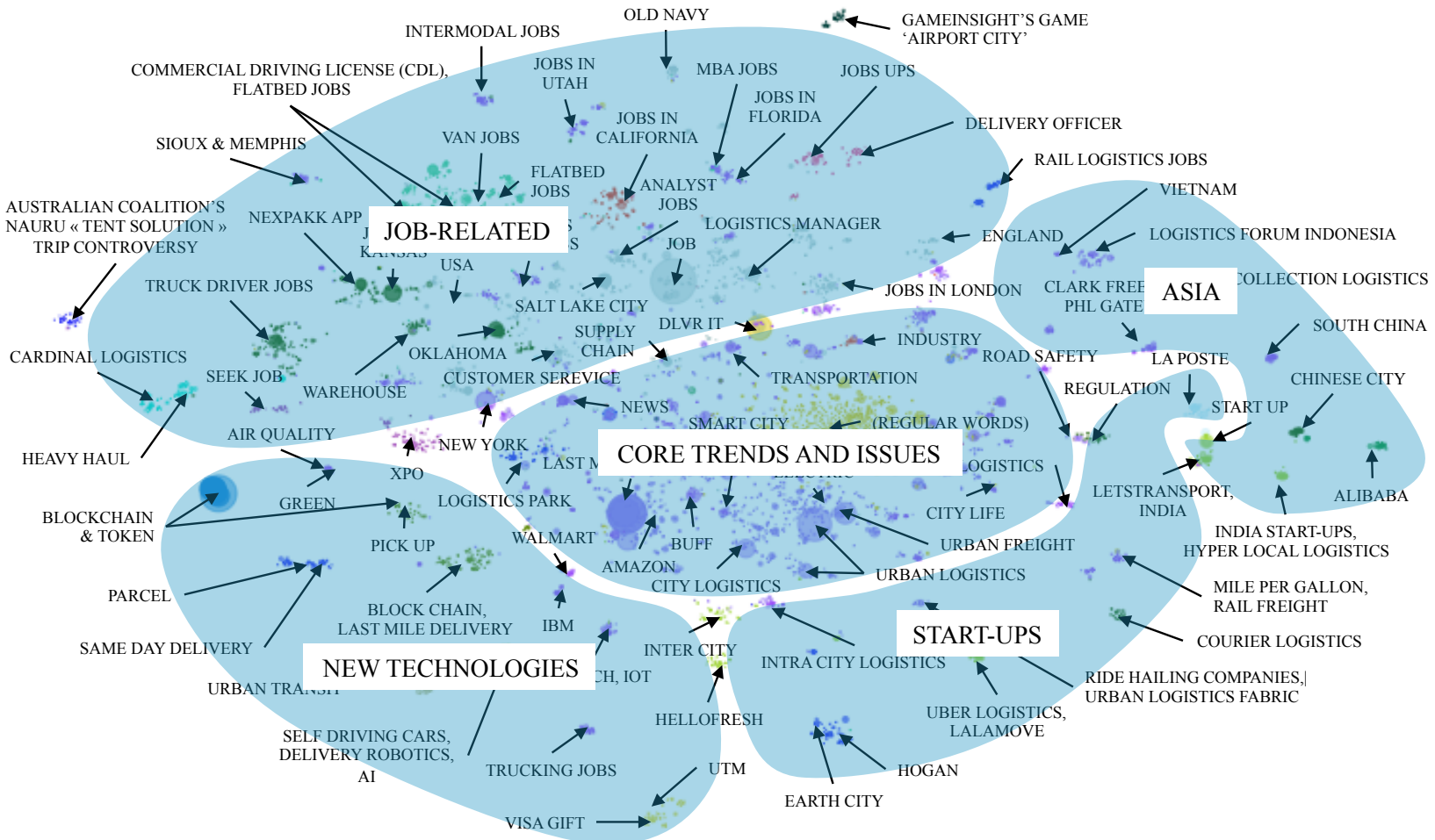

Figure 5. Main clusters in the interest map

Job offers occupy a big share of the corpus. In comparison, regulation, for example, is much less prevalent. This makes sense given the nature of Twitter and its usage: the job market requires advertising positions and given the mere size of the market and its turnover, a substantial throughput is expected. An open question is whether one can draw useful insights, such as analysing tensions on the market, or anticipating future developments.

Many tweets address new technologies, start-ups or innovative firms. This also makes sense, as very often these technologies or firms require visibility to build their image and raise funds. Communication on Twitter probably contributes efficiently to these objectives. Incidentally, for researchers and policymakers, social media mining can be a very cost-efficient business intelligence process, especially in such a fast-changing environment.

In order to assess what social media mining can bring to the observation of City Logistics, it is critical to identify under-represented issues and/or blind spots. For one, regulation and policy issues are present, but not easily visible. Quite satisfyingly, one can find a rather large range of issues (e.g. road safety, fuel consumption, sustainability, urban fabric, etc.) and solutions (e.g. training, ICT, urban consolidation centres, clean vehicles, cargo-bikes, etc.). However, one should not take prevalence as an index of importance, one way or another. In contrast, some concepts, very much advertised in academic circles, are almost absent in the corpus (e.g. 21 tweets about the Physical Internet, 8 about off-hour deliveries, 3 about synchro-modality). Particularly striking given the nature of the corpus is the *virtual absence* of issues such as labour regulation, or negative local impacts of urban freight (pollution, noise, etc.). It is possible that the corresponding stakeholders are vocal on other forms of social media or use other keywords than those used in our query. Finally, our methodology does not allow an easy identification of stakeholders nor their characteristics or issues, let alone their strategies. On this topic, among others, expertise is still needed.

The corpus can also be analysed with a dynamic approach. For example, one can examine the evolution of the prevalence of a topic, in combination with sentiment analysis. Figure 4 presented the evolution of all City Logistics tweets over the last ten years. This analysis can be broken down into more specific topics, as illustrated by Figure 6.

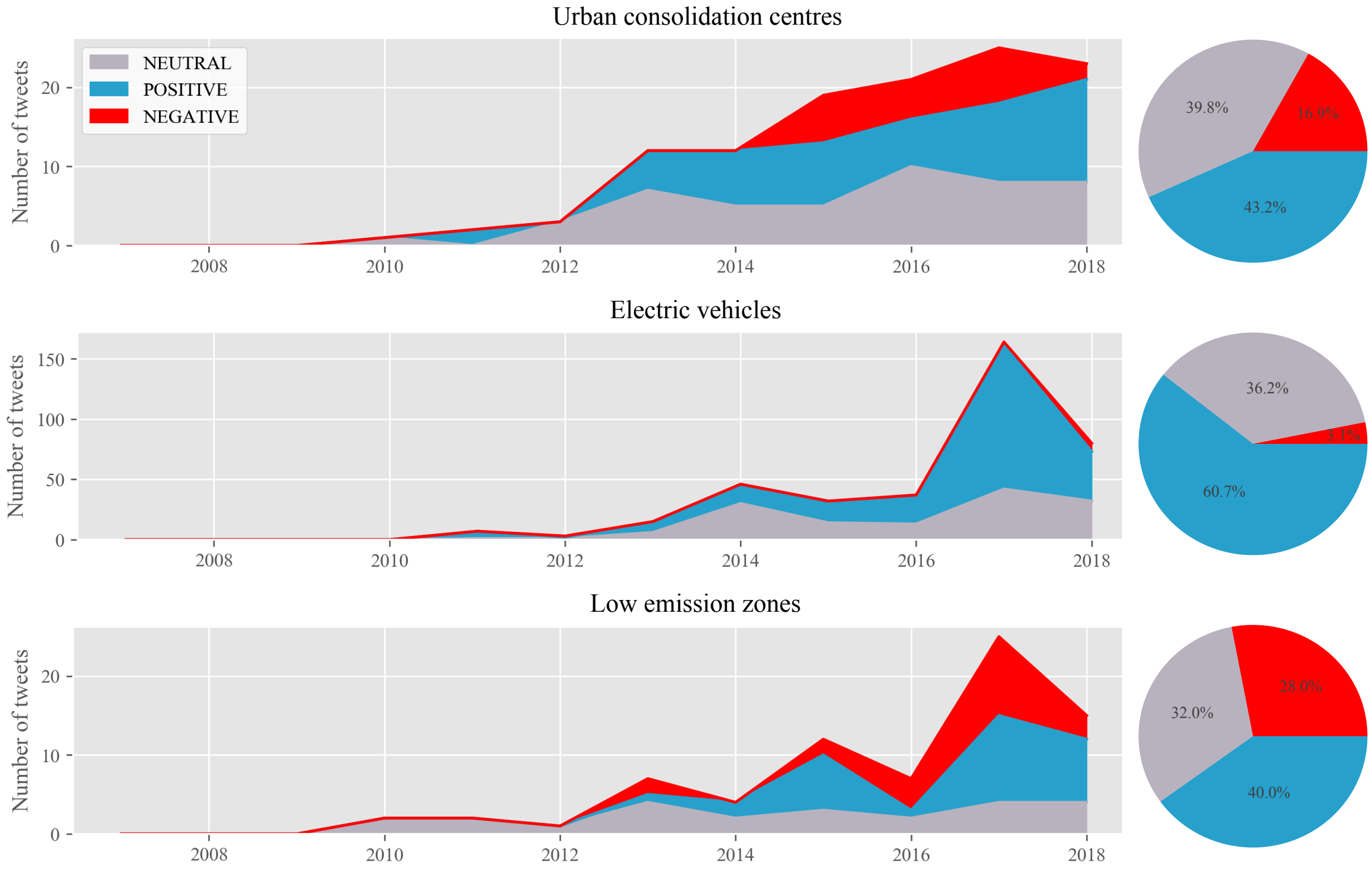

Figure 6. Sentiment distribution and evolution of specific topics

The three topics illustrated in Figure 6 were chosen due to their varied dynamics. The first topic, *urban consolidation centres*, is present early in the corpus, and its dynamic is little more than stable (its relative prevalence is receding). Also, opinions shifted from positive to mixed, with a more significant proportion of negative perceptions in 2015. This result ratifies the validity of our approach, as during this period many of these centres closed due to lack of financial viability. The two other topics—electric vehicles and low emission zones—are more dynamic, but sentiments differ: while the topic of electric vehicles is consistently associated with positive messages, sentiments about low emission zones are not as consensual.

## CONCLUSION

This paper performed social media mining about City Logistics using Twitter. The proposed methodology used Machine Learning and Natural Language Processing tools on a corpus of 111 265 Twitter entries. Two main contributions were presented: (1) An interactive interest map, which allows the display of concepts that are noteworthy for people who tweet about City Logistics. This map allows the visualisation of concepts that are more or less significant (in terms of frequency of appearance) and it shows proximity between those concepts. (2) Sentiment analysis was performed on the collected data. This allowed us to assess that the overall view of City Logistics is more positive than negative, as the sentiment distribution of the corpus is 48% neutral, 45% positive and 7% negative.

Statistical analysis of the most prevalent n-grams in the corpus brings us to the conclusion that the preferred term to use in reference to our subject is "*City Logistics"* as opposed to *Last-Mile Logistics, Urban Logistics* or *Urban Freight.* This analysis also shows that the most important topic in the City Logistics tweets is Employment. The interest map reveals distinctive clusters such as *employment* (job offers), *new technologies* (self-driving cars, blockchain, IoT) and *start-ups and new forms of organization* (ride hailing, courier logistics, hyper-local logistics). However, it is to note that in the centre of the map (central cluster) we find issues related to *quality of life, zero emissions, regulation, smart city, etc*. It is important to highlight that the large number of tweets related to *employment* reveals that the corpus is biased. We must bear in mind that many Twitter users work in institutional communication (their job is to create social media content). As a result, caution is binding when interpreting these results, as the analysis carried out in this paper does not generalize the vision of the general population about City Logistics. It only reveals the view of Twitter users.

This exploratory research could only scratch the surface of the topic. A clear strength of social media analysis is how it can cost-efficiently contribute to business and technological intelligence, with the risk, however, of missing less-advertised topics. Regarding dynamics and sentiment analysis, it seems that there is untapped potential; this clearly requires more work. With respect to public policy issues, the picture is less favourable: several topics are present, but they are not prominent; and some of them are virtually non-existent, in striking contrast with other environments.

Social media constitutes an opportunity to complete or even partially replace classic observation protocols. At first sight, they are a formidable opportunity: massive data in natural language coming from many people, with messages that can potentially be located, dated and linked to users. However, this opportunity must be verified: are all opinions expressed without biases? Are some stakeholders more vocal than others? How reliable is the information? Are some trends over-represented while other are–deliberately or not–understated? These questions (and probably many others) need answers before social media can be considered a reliable protocol for the observation of City Logistics, i.e. a protocol in which the biases and blind spots are correctly identified.

City Logistics is a complex system; it is at the same time political, economic, social and technological. Despite its potential, but perhaps not surprisingly, social media mining cannot provide a complete and understandable picture of City Logistics with all these dimensions (at least not with our methodology). Human expertise will still be required, for a while.

## ACKNOWLEDGEMENT

This work is supported by the Urban Logistics Chair at MINES ParisTech PSL, sponsored by ADEME (French environment and energy management agency), La Poste, Marie de Paris (City of Paris), Pomona Group and RENAULT.